\pdfoutput=1
\documentclass[10pt,twocolumn,letterpaper]{article}

\usepackage{iccv}
\usepackage{times}
\usepackage{epsfig}
\usepackage{graphicx}
\usepackage{algorithm}
\usepackage{algorithmic}
\usepackage{amsmath}
\usepackage{amssymb}

\usepackage[pagebackref=true,breaklinks=true,letterpaper=true,colorlinks,bookmarks=false]{hyperref}

\iccvfinalcopy 

\def\iccvPaperID{****} 

\ificcvfinal\fi

\usepackage{caption}
\usepackage{booktabs}

\begin{document}

\title{Feature Pyramid Network for Multi-task Affective Analysis}

\author{Ruian He\thanks{Equal Contribution},\quad  Zhen Xing\footnotemark[1],\quad Weimin Tan,\quad Bo Yan\\
School of Computer Science, \\Shanghai Key Laboratory of Intelligent Information Processing, Fudan University\\
{\tt\small \{rahe16, xingz20, wmtan14, byan\}@fudan.edu.cn}
}

\maketitle
\ificcvfinal\thispagestyle{empty}\fi

\begin{abstract}
   Affective Analysis is not a single task, and the valence-arousal value, expression class, and action unit can be predicted at the same time. Previous researches did not pay enough attention to the entanglement and hierarchical relation of these three facial attributes. We propose a novel model named feature pyramid networks for multi-task affect analysis. The hierarchical features are extracted to predict three labels and we apply a teacher-student training strategy to learn from pretrained single-task models. Extensive experiment results demonstrate the proposed model outperforms other models. This is a submission to The 2nd Workshop and Competition on Affective Behavior Analysis in the wild (ABAW). The code and model are available for research purposes at \href{https://github.com/ryanhe312/ABAW2-FPNMAA}{this link}.
\end{abstract}

\section{Introduction}

Numerous researches have been conducted on affective analysis for years and aim to automatically comprehend human feelings, emotions, and behaviors. Affection analysis contributes to many applications, such as psychological therapy and marketing analysis. 

We participate in the 2nd Workshop and Competition on Affective Behavior Analysis in-the-wild (ABAW).\cite{kollias2021analysing}\cite{kollias2019deep}\cite{kollias2019expression}\cite{kollias2019face}\cite{kollias2020analysing}\cite{kollias2021affect}\cite{kollias2021distribution}\cite{zafeiriou2017aff} And there are three challenges in affective analysis:  valence-arousal, seven basic expression and action unit prediction. Valence-arousal value is a description of human expression, and there are positive and negative for valence and high and low in arousal. Seven basic expressions\cite{kollias2021analysing} are defined as Anger, Disgust, Fear, Happiness, Sadness, Surprise, and Neutral here. All facial movement can be categorized in different action units in terms of the Facial Action Coding System (FACS) model\cite{Ekman1978FacialAC}. 

The model we propose is named feature pyramid networks for multi-task affective analysis. It is an image-based method and can solve the three tasks above at the same time. And we use Feature Pyramid Network\cite{Lin2017FeaturePN} for the backbone and the connection between feature layers can bring feature fusion merits. And we will cover this in section \ref{sec/fpn}.

We use the cropped and aligned images from Aff-Wild2 dataset provided by ABAW2 as the training and validation set. There are a great number of missing labels in the provided AffWild2 dataset. Many samples only have labels for one or two tasks. So we augment the dataset with ExpW(Expression-in-the-Wild)\cite{Zhang2017FromFE} and AffectNet\cite{Mollahosseini2019AffectNetAD} and follow the method in semi-supervised learning and propose a teacher-student training procedure to train a multi-task model. We will get into details in section \ref{sec/teacher}.

\begin{figure}[!t]
\centering
\includegraphics[width=0.45\textwidth]{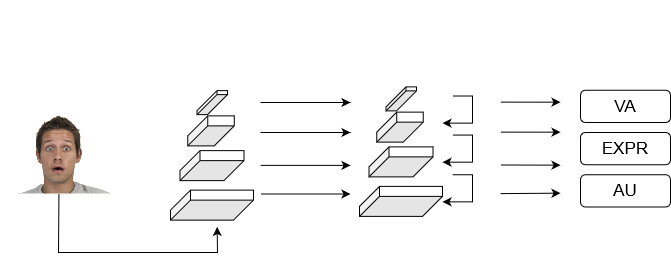}
\caption{Multi-task Feature Pyramid Network. We use the backbone of feature pyramid network followed by three classification heads as our model. The features extracted in each FPN layer are concatenated and fed to the classifier. The human image is from ExpW.\cite{Zhang2017FromFE}}
\label{img/architecture}
\end{figure}

Finally, our contributions can be concluded as follows:

1. We propose a novel model using the feature pyramid network for multi-task affective analysis. It exploits the entanglement and hierarchical relation of these three facial attributes.

2. We apply a teacher-student training procedure to learn from missing labels. This allows the multi-task model to learn from all labels in the Aff-Wild2 dataset at the same time.

3. We achieve an excellent result on the Aff-Wild2 validation set over the official baseline\cite{kollias2021analysing} and other models.

\section{Related Works}

Typically, according to the type of extracted features, FER algorithms can be categorized into hand-crafted feature-based and deep learning-based methods. Hand-crafted features can be further divided into texture-based features, geometry-based features, and hybrid features. The texture-based features mainly contain Gabor \cite{2002Gabor}, SIFT \cite{2003SIFT}, HOG \cite{2005Histograms}, LBP \cite{2011}, NMF \cite{2011Graph}, etc. The geometry-based features are primarily based on facial landmark points. Hybrid features refer to the combination of the two above-mentioned features. In the past few years, large-scale facial expression datasets, such as RAF-DB \cite{2017Reliable} and AffectNet \cite{Mollahosseini2019AffectNetAD}, greatly facilitate deep FER research. Very Recently, Zeng \textit{et al.} \cite{DBLP:conf/eccv/ZengSC18} first considered the inconsistent annotation problem in FER research. Wang \textit{et al.} \cite{DBLP:conf/cvpr/WangPYL020} suppressed the uncertainties and improved the FER performance. Li \textit{et al.} \cite{DBLP:journals/tip/LiWDYG21} considered the long-tail distribution and high similarity in FER datasets and achieved leading performance.

The task of facial action unit detection is to recognize active action units. Generally, action units detection methods can be described into two groups: patch-based methods and structure-based methods. Patch-based methods target to extract local features from important facial regions. Zhao \textit{et al.} \cite{DBLP:conf/cvpr/ZhaoCZ16} proposed a region layer to induce important facial regions for 
extracting local features. In \cite{DBLP:conf/fgr/LiAZY17}\cite{DBLP:journals/pami/LiAZY18}, Li defined regions of interest (ROI) according to facial landmark points and designed individual convolutional layers to learn deeper features for each facial region. Shao \textit{et al.} \cite{DBLP:conf/eccv/ShaoLCM18} applied multi-scale region learning to extract AU-related local features. Recently, structure-based methods \cite{DBLP:conf/cvpr/NiuHYHS19}\cite{DBLP:conf/aaai/FanLL20}\cite{DBLP:conf/aaai/SongCZJ21} tend to use CNNs and GCNs to extract the local information and capture the relationship between AUs respectively.

Multi-task learning in computer vision is aimed at improving the generalization ability of network on related tasks, e.g. surface normal estimation and edge labels \cite{DBLP:conf/cvpr/WangFG15}, camera pose estimation and wide baseline matching\cite{DBLP:conf/eccv/ZamirWAWMS16}, person attribute classification\cite{DBLP:conf/cvpr/LuKZCJF17} etc. MTL networks can also be divided into two categories. The methods of first category target to find which part of the baseline should be shared and which part should be task-specific \cite{2015Learning}\cite{DBLP:conf/iclr/MeyersonM18}\cite{DBLP:conf/cvpr/MisraSGH16}\cite{DBLP:conf/aaai/RuderBAS19}\cite{DBLP:conf/iccv/BragmanTOAC19}, while the methods of second category make task grouping by taking task-similarity into account \cite{DBLP:journals/jmlr/XueLCK07}\cite{DBLP:conf/nips/JacobBV08}\cite{DBLP:conf/icml/KangGS11}\cite{DBLP:conf/cvpr/LuKZCJF17}\cite{DBLP:conf/cvpr/MejjatiCK18}.

\section{Multi-task Feature Pyramid Network}

\label{sec/fpn}

Valence-arousal, expression, and facial action units are not all about extracted top features. The low-level features which contain local information can also help with classification. So we exploit the pyramid network to predict affective labels.


Since the three tasks all describe affective behaviors, they are closely related. Specifically, expression categories and valence-arousal describe affective behaviors globally, while action units reflect the local facial movements. For example, the occurrence of AU6 and AU12 is a sign of happiness \cite{1983EMFACS}, while happy images always have high valence scores\cite{Mollahosseini2019AffectNetAD}.  

There are a few earlier works concerning the use of feature pyramid in facial expression recognition\cite{Yang2021ACF}\cite{Su2021FacialER}.  However, they only focus on a single task and we show that the feature pyramid also works for multi-task affective analysis.

\subsection{Network Architecture}

Our network architecture can be displayed as figure \ref{img/architecture}. We follow the implementation of Feature Pyramid Networks to build our model\cite{Lin2017FeaturePN}. The input image is feed into the ResNet \cite{He2016DeepRL} backbone first and extract conv2, conv3, conv4, and conv5 output. Then the top-down feature fusion is applied. The up-layer features are scaled by 2 and added to the down-layer. We do average pooling in all four layers to get $1\times 1$ output per layer. Finally, features from all layers are concatenated to form a vector used for classification.

\subsection{Multi-task learning}

We use three different classification heads for the three tasks respectively. For the valence-arousal task, we use a linear layer and mean squared error(MSE) loss function. The valence-arousal task is a regression task and it is expected to predict values between -1 and 1 for valence and arousal intensity. Compared to mean absolute error(MAE), MSE has more tolerance for small deviation which is good for being not overfitting. The loss can be calculated as:

\begin{equation}
    L_{VA}(x,y) = (x_v - y_v)^2 + (x_a - y_a)^2
\end{equation}
in which $x_v$ and $y_v$ stand for the prediction and ground truth of valence and $x_a$ and $y_a$ stand for the prediction and ground truth of arousal.

For the expression task, we use the cross-entropy loss function for this multi-class prediction. And for the facial action unit task, we use binary cross-entropy loss function for this multi-label prediction. It is a typical solution for expression classification. The CE and BCE loss can be calculated as:

\begin{equation}
    L_{EXPR}(x,y) = -x[y] + \log(\sum_j\exp(x[j]))
\end{equation}

\begin{align}
        &L_{AU}(x,y) =\\ \nonumber 
        &- \sum_j(y_j\log(\sigma(x_j))+(1-y_j)\log(\sigma(1-x_j)))
\end{align}
In cross-entropy loss, $y$ is the class number and in binary cross-entropy loss, $y$ is the 0-1 binary vector.

Our total loss function for the multi-task model can be shown as:

\begin{equation}
    L_{Multi} = \alpha L_{VA} + \beta L_{EXPR} + \gamma L_{AU}
\end{equation}
And we finally choose $ \alpha=\beta=\gamma=1$ for training. All three classification heads only have one linear layer and share the same backbone. We expect the backbone network to learn unified distinguishable features of expression. And three heads get features from all four layers in FPN.

\section{Learning from Missing and Unbalanced Labels}

\label{sec/teacher}

\subsection{Data Augmentation}

Aff-Wild2, the dataset we mainly use for the ABAW2 Competition, is annotated for three affective behavior analysis tasks: valence-arousal estimation, basic expression classification, and facial action unit detection. 

\begin{figure}[!t]
    \centering
    \includegraphics[width=\columnwidth]{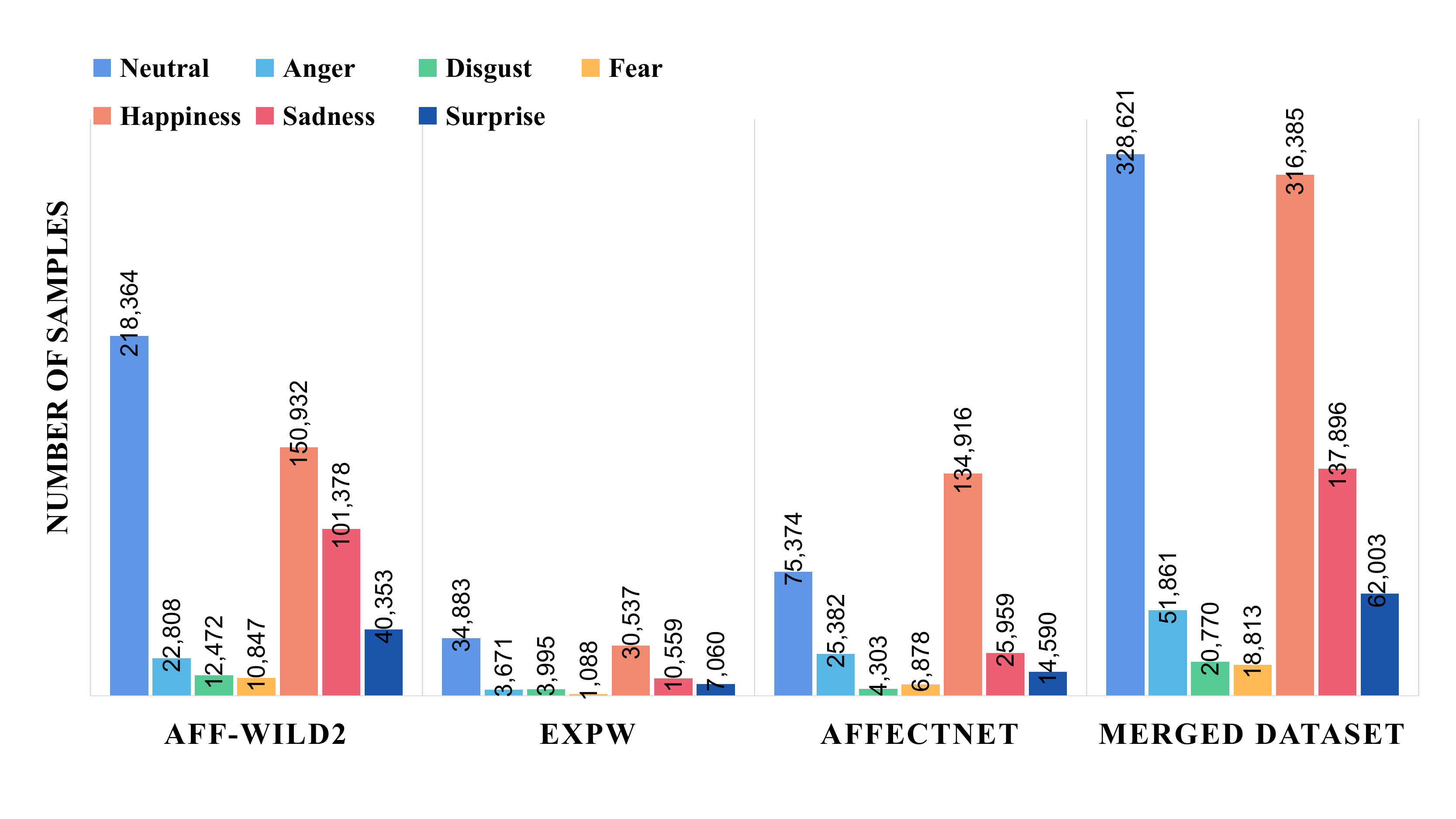}
    \caption{Comparison of different basic expression distributions. It shows the label distribution of the three datasets.}
    \label{fig:expr_dist}
\end{figure}

For expression classification, we can observe a severe imbalance distribution in figure \ref{fig:expr_dist}. Re-balancing is a frequently-used strategy to deal with long-tail distribution, but it will cause under-fitting to the head or over-fitting to the tail. So we just merge samples of Aff-Wild2, ExpW \cite{Zhang2017FromFE} and AffectNet \cite{Mollahosseini2019AffectNetAD} without re-sampling. The distributions of seven emotion categories in the Aff-wild2 dataset, the ExpW dataset, the AffectNet dataset, and the merged dataset are shown in figure \ref{fig:expr_dist}.

For valence-arousal estimation, we import the AffectNet dataset which contains 280,000 images annotated with valence-arousal scores in [-1, 1]. Cause we regard it as a regression problem, we do not apply any re-balancing strategies. 

\begin{figure}[!t]
\centering
\includegraphics[width=0.3\textwidth]{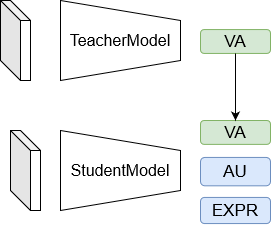}
\caption{Teacher-student Training Strategy. The teacher model is a single task model and the student model is a multi-task model. In this situation, the VA label is missing and our single-task model provides the soft label for the multi-task model. So the multi-task model can learn from all three labels simultaneously.}
\label{img/training}
\end{figure}

\begin{table*}[ht]
\centering
\caption{Experiments on AffWild2 Validation Set. VA, EXPR, and AU mean the score for each task. CCC, F1, Acc, and score follow the rule that the higher the better. For different input data, Single is the origin AffWild2 dataset. Mixed is our augmented dataset. And Multi is the AffWild2 dataset with our generated soft labels.}
\begin{tabular}{@{}l|l|lll|lll|lll@{}}
\toprule
Model     & Data     & CCC-A & CCC-V & VA    & Acc    &  F1  & EXPR & Acc   &  F1 & AU   \\ \midrule
Baseline\cite{kollias2021analysing}  & Single & 0.23  & 0.21  & 0.22  & 0.50   & 0.30   & 0.36 & 0.22 & 0.4  & 0.31 \\ \midrule
MobileNet & Single & 0.34  & 0.12  & 0.23  & 0.50   & 0.29  & 0.36 & 0.87 & 0.39 & 0.63 \\
ResNet    & Single & 0.29  & 0.14  & 0.22  & 0.52  & 0.30   & 0.38 & 0.86 & 0.41 & 0.64 \\
F3R(Ours)       & Single & 0.27  & 0.22  & 0.24  & 0.55  & 0.31  & 0.39 & 0.87 & 0.43 & 0.65 \\ \midrule
MobileNet & Mixed    & 0.31  & 0.27  & 0.29  & 0.58  & 0.36  & 0.43 & 0.87 & 0.37 & 0.62 \\
ResNet    & Mixed    & 0.28  & 0.23  & 0.26  & 0.57 & 0.36 & 0.43 & 0.87 & 0.41 & 0.64 \\
F3R(Ours)        & Mixed    & 0.34 & 0.31 & 0.32 & 0.59  & 0.37  & 0.44 & 0.87 & 0.37 & 0.64 \\ \midrule
F3R(Ours)        & Multi    & 0.44  & 0.28  & 0.36  & 0.61   & 0.40   & 0.47 & 0.88 & 0.40  & 0.64 \\ \bottomrule
\end{tabular}
\label{tab/result}
\end{table*}

\subsection{Training Strategies}

We use teacher-student network to fix missing labels. The similar work is \cite{Shin2020SemisupervisedLW}, and we also use  multi-teacher-single-student(MTSS) strategy but in a simpler way. The
process can be displayed as figure \ref{img/training}. The  teacher model is a single-task model and student model is a multi-task  model. We don't match the feature space but just use the single-task teacher model to predict the missing label. The multi-task student network shares the same structure of the teacher model but has several classification/regression heads to learn general features for all three expression recognition at the same time. And the process can also be showed as the following algorithm:

\begin{algorithm} 
	\caption{Teacher-student Training Strategy} 
	\label{alg:teacher} 
	\begin{algorithmic}
		\STATE run preprocessing scripts for mixing AffWild2, ExpW and AffectNet datasets. The dataset $D_{mixed}$ is different for each task.
		\FOR{data type $t \in \{VA,EXPR,AU\}$ }
        \STATE train single teacher model $\phi_t$ for each task.
        \STATE predict label $L_t$ for every image in AffWild2 dataset.
        \ENDFOR
        \STATE build unified AffWild2 dataset $D_{multi}$ from predicted labels $L_t , t \in \{VA,EXPR,AU\}$.
        \STATE train student multi-task model $\phi_{multi}$ in $D_{multi}$.
        \RETURN $\phi_{multi}$
	\end{algorithmic} 
\end{algorithm}

We limit the batch of each epoch to 25\% to avoid overfitting. As the AffWild2 dataset is combined with video frames, there are a lot of redundant labels for each task. The differences between frames are not obvious enough to provide more information to train the model. So we reduce the steps of each epoch to 25\% of the whole training set. That is to say in each epoch we randomly get 25\% samples of the training set. And that makes the training process more smooth and robust.

The model network is trained on an NVIDIA RTX2080 GPU. For a single-task model, the network is trained for 40 limited epochs and it cost 3 hours. And for the multi-task model, it cost 1 day for training 20 full epochs. We use
Adam optimizer\cite{Kingma2015AdamAM} at a learning rate of 1e-3. And the batch size is set to 128.

\section{Experiment}

We have done extensive experiments on the validation set of the AffWild2\cite{kollias2021analysing}\cite{kollias2019deep}\cite{kollias2019expression}\cite{kollias2019face}\cite{kollias2020analysing}\cite{kollias2021affect}\cite{kollias2021distribution}\cite{zafeiriou2017aff} . We have compared our model with typical classification models such as MobileNet\cite{Howard2017MobileNetsEC} and ResNet \cite{He2016DeepRL}.  We use pretrained model of MobileNet and ResNet and FPN for both student model and teacher model. We also refer to the official baseline on the top, which is taken from the paper\cite{kollias2021analysing} . 

We have also tested on different data settings. Firstly, we only train on the original dataset and then on our augmented dataset. At last, we use the teacher-student training strategy to train a multi-task model. The quantitative results can be found in table \ref{tab/result}.

\subsection{Metrics}

For evaluation, we follow the settings in the baseline paper \cite{kollias2021analysing}. The metrics are CCC, F1, and TAcc. The Concordance Correlation Coefficient(CCC) evaluates the agreement between two time series by scaling their correlation coefficient with their mean square difference. We use CCC to evaluate the prediction of the valence-arousal task. The CCC can be calculated as:

\begin{equation}
    CCC = \frac{2s_{xy}}{s_x^2+s_y^2+(\overline{x}-\overline{y})^2}
\end{equation}

in which $s_{xy}$ is the covariance of prediction and ground truth and $s_x$ and $s_y$ are the separated variance.

The $F_1$ score is a weighted average of the recall (i.e., the ability of the classifier to find all the positive samples) and precision (i.e., the ability of the classifier not to label as positive a sample that is negative).$F_1$ score is used to evaluate the prediction of seven basic expression task. 

\begin{equation}
    F_1 = \frac{2\times precision \times recall}{precision + recall}
\end{equation}

And the average $F_1$ is denoted as $AF_1$, which means the average of all label predictions. And $AF_1$ is used to evaluate the task of action unit prediction.

Total accuracy (denoted as TAcc) is another metric for classification and it is defined on all test samples and is the fraction of predictions that the model got right. This is also used for expression and action unit tasks. 

\begin{equation}
    TAcc = \frac{Correct\ Predictions}{All\ Predictions}
\end{equation}

The three tasks scores are calculated as follows, and they are the weighted sum of above metrics.
\begin{align}
    &S_{VA} = 0.5\times CCC_{V} + 0.5\times CCC_{A} \\
    &S_{EXPR} = 0.67\times F_1 + 0.33\times TAcc \\
    &S_{VA} = 0.5\times AF_1 + 0.5\times TAcc 
\end{align}

\begin{figure}[!t]
    \centering
    \includegraphics[width=\columnwidth]{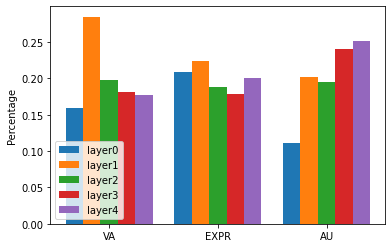}
    \caption{Feature selection visualization of classification head. The layer 0 to 4 is extracted from our multitask FPN network and the features are layer0 are low level features while the features in layer4 are high level features. This figure show how these layers contribute to the classification.}
    \label{fig:index}
\end{figure}

\subsection{Results}

From the table \ref{tab/result} we can see that data augmentation helps with almost every model. Firstly, we have done experiments on the bare AffWild2 dataset with our models. The result was mediocre. Then we supplemented the dataset with ExpW\cite{Zhang2017FromFE} and AffectNet\cite{Mollahosseini2019AffectNetAD} to make up for its lack of diversity. The data augmentation improved valence-arousal and expression prediction scores by approximately 10\%.

And we used the best single-task model to generate a unified AffWild2 dataset for multi-task training. We applied a teacher-student training strategy to train these models and all of them(Multi) are better than the single-task one(Single) in the bare AffWild2 dataset.

Compared to other models like MobileNet\cite{Howard2017MobileNetsEC} and ResNet \cite{He2016DeepRL}, our feature pyramid networks for multi-task affective analysis model outperforms by 2\% to 10\%. Though the improvement is not as significant as manipulating the data, it is still better than ResNet which has the same amount of parameters.

We also visualize the contribution of each layer to the classification in our multitask model as figure \ref{fig:index}. We sum all the absolute weight in the classification head connected to the specific layer and multiply it with the relative size of the input tensor. The figure shows that every task has a preference on feature selection, e.g. VA relies on layer1 and AU relies on the last two layers. Our backbone provides the hierarchical features and the classification head learns to choose critical features.

\section{Conclusion}

We propose a novel model using the feature pyramid network for multi-task affective analysis. The model exploits the hierarchical features in the backbone network and makes predictions for three different tasks. This model outperforms other classification models and the official baseline. 

We also use a teacher-student training strategy to learn from missing labels in the dataset. It allows the model to learn from all three labels simultaneously. Without this strategy, multi-task learning won't take effect. 

Though the model has an excellent result, the mechanism behind is not quite studied. The feature selection in the classification head is not clear and needs further researches. And this model is not robust for side faces and occluded faces. More facial prior should be added to have a full understanding of the human face.

{\small
\bibliographystyle{ieee_fullname}
\bibliography{main}
}

\end{document}